\documentclass{article}

\usepackage{microtype}
\usepackage{subcaption,graphicx}
\usepackage{booktabs} 
\usepackage{tabu}
\usepackage{breakcites}
\usepackage{multirow}
\usepackage[a4paper]{geometry}
\usepackage[table,xcdraw]{xcolor}

\usepackage{hyperref}
\hypersetup{
    colorlinks=true,
    linkcolor=blue,
    citecolor = blue
}

\usepackage{natbib}
\usepackage{amsmath}
\usepackage{amssymb}
\usepackage{mathtools}
\usepackage{amsthm}
\usepackage{tikz}
\usetikzlibrary{cd}
\usepackage{adjustbox}
\newcommand{\correction}{}

\usepackage{authblk}
\usepackage{float}

\usepackage[capitalize,noabbrev]{cleveref}

\theoremstyle{plain}

\theoremstyle{definition}

\theoremstyle{remark}

\usepackage[textsize=tiny]{todonotes}

\title{Interpretability of Graph Neural Networks to Assess Effects of Global Change Drivers on Ecological Networks}
\author[1]{Emre Anakok}
\author[2]{Pierre Barbillon}
\author[3]{Colin Fontaine}
\author[4]{Elisa Thebault}
\affil[1,2]{Université Paris-Saclay, AgroParisTech, INRAE, UMR MIA Paris-Saclay, 91120, Palaiseau, France.}
\affil[3]{Centre d'Écologie et des Sciences de la Conservation, MNHN, CNRS, SU, 43 rue Buffon, 75005 Paris, France}
\affil[4]{Sorbonne Université, CNRS, IRD, INRAE, Université Paris Est Créteil, Université Paris Cité, Institute of Ecology and Environmental Sciences (iEES-Paris), 75005 Paris, France}
\affil[1]{emre.anakok@u-paris.fr}

\date{}                     
\begin{document}

\maketitle

\begin{abstract} 
Pollinators play a crucial role for plant reproduction, either in natural ecosystem or in human-modified landscape. Global change drivers,including climate change or land use modifications, can alter the plant-pollinator interactions. 
 To assess the potential influence of global change drivers on pollination, large-scale interactions, climate and land use data are required. While recent machine learning methods, such as graph neural networks (GNNs), allow the analysis of such datasets, interpreting their results can be challenging. We explore existing methods for interpreting GNNs in order to highlight the effects of various environmental covariates on pollination network connectivity. \correction{An extensive} simulation study is performed to confirm whether these methods can detect the interactive effect between a covariate and a genus of plant on connectivity, and whether the application of debiasing techniques influences the estimation of these effects. An application on the Spipoll dataset, with and without accounting for sampling effects, highlights the potential impact of land use on network connectivity and shows that accounting for sampling effects partially alters the estimation of these effects.

\end{abstract}

\includegraphics[width= \textwidth]{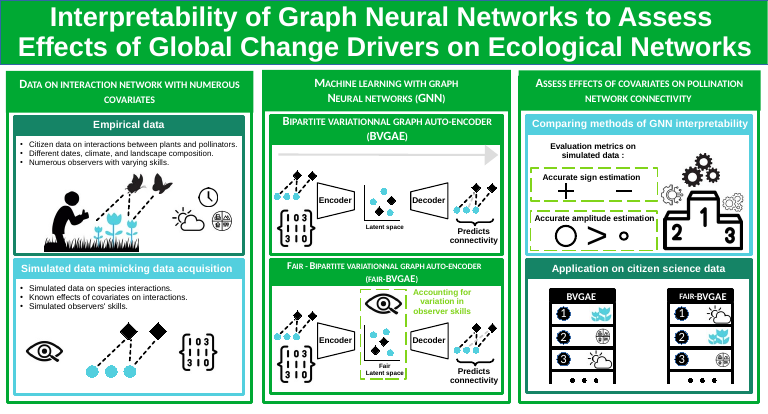}

\section{Introduction}

Pollinators are affected by pressures related to human impact on the environment such as climate warming and land use, as several studies have reported their population declines \citep{bartomeus_historical_2018,imperatriz-fonseca_assessment_2016}.

Interactions between plants and pollinators can be studied using bipartite networks \citep{ings_review_2009}, with nodes representing plant and pollinator taxa, while edges denote observed interactions. 

Analyzing the structure of plant-pollinator interaction networks has been insightful for understanding how pollinators respond to human pressures. Research has shown that land use has a strong effect on pollinator composition, abundance, and network structure. More specifically, studies have identified that agricultural land cover, in contrast to urban uses, could increase pollinator generality, as well as their robustness to extinction scenarios \citep{deguinesWhereaboutsFlowerVisitors2012,redhead_potential_2018}. However, agricultural intensification is likely a main contributor of the decline of wild pollinator species \citep{duchenne_longterm_2020}. Urban settings tend to be associated with lower pollinator biodiversity compared to agricultural land uses, but pollinator abundance can vary significantly depending on specific urban land uses such as gardens, allotment, parks or artificial surfaces \citep{baldock_systems_2019}, as well as flowers that are present \citep{baldock_systems_2019,rollings_quantifying_2019}. As for climate warming, it has been associated with earlier mean flight date for pollinators \citep{duchenne_phenological_2019}, and could benefit some bee species \citep{duchenne_longterm_2020}.

Large-scale interaction data are necessary to assess the potential influence of land use and climate conditions on ecological networks. Two approaches can be used to study this influence. First, distinct networks across different conditional or environmental settings can be compared. This method yields interesting results about the effects on ecological networks of seasons \citep{fisogni_seasonal_2022}, urbanism \citep{fisogni_seasonal_2022,dore_relative_2021} or altitude \citep{Lara_Romero_2019}. An alternative approach could be to aggregate data from multiple conditions into a single unified network.  Species co-occurrences observed under different conditions are used to construct multiple networks, with connections inferred from the unified network. This approach has uncovered correlation between agricultural land cover and both pollinator generality and robustness to extinctions \citep{redhead_potential_2018}, or between land use and food web structure \citep{tetrapod}.  

The accumulation of interaction data at a very large scale has been facilitated by citizen science programs, notably the French program Spipoll \citep{deguinesWhereaboutsFlowerVisitors2012} that monitors plant-pollinator interactions across metropolitan France since 2010. Participants are asked to take pictures of pollinators visiting a freely choosen flowering plant during a 20-minutes observation session, then upload the pictures and identify the insects on a designated website. Each session contains the set of insects observed on a plant species at a given time and location. With around 500,000 plant-pollinator interactions recorded, the date and place of observations enabled the extraction of corresponding climatic conditions, from the European Copernicus Climate data set \citep{cornesEnsembleVersionEOBS2018}, and the corresponding land use proportion from the Corine Land Cover (CLC, \citealt{CLC2018}), with 44 categories in a 1000m radius around the observation location.

However, citizen science programs are prone to sampling bias due to the multiplication of observers and associated observer effect \citep{jiguetMethodLearningCaused2009,birdStatisticalSolutionsError2014,kellingCanObservationSkills2015,johnstonEstimatesObserverExpertise2018}. For the Spipoll dataset, sampling bias has been observed due to differences between users \citep{deguinesFunctionalHomogenizationFlower2016} and the accumulation of experience by each user \citep{deguinesFosteringCloseEncounters2018}.

The vast amount of interaction data, along with climatic and land use data can be jointly analyzed using recent developments in machine learning such as graph neural networks (GNNs), which can also account for sampling bias \citep{anakok2024BFGVAE}. GNNs have demonstrated improving performance on various artificial intelligence tasks on networks and are increasingly popular. However, GNNs\correction{, as neural networks in general,} often operate as ``black boxes'', making their outputs hardly interpretable despite the high accuracy of predictions. \correction{To popularize these  methods in ecology, there is a need to obtain interpretability \citep{Cipriano}. } Methodologies have been developed to interpret neural networks \citep{zhang_survey_2021,fanInterpretabilityArtificialNeural2021}. \correction{Such approaches have been applied in ecology \citep{scrinzi} for  updating forest inventories}. Graph neural networks especially require specific adaptation due to the discrete nature of graph data \citep{wu_interpretability_2022,yuan_explainability_2022,khan_interpretability_2023}.

Without access to ground truth, one can rely on simulations mimicking expected ecological processes to validate the methodological approaches, as some behaviors are expected to be captured by such GNN interpretability methods. Notably, they should be able to detect (i) preferential relationships between plants and pollinators, (ii) the influence of environmental covariates, whether positive or negative, and (iii) potential interactive effects between covariates and plant and/or pollinator identity, i.e.~a covariate affecting differently the probability of interaction of two plant-pollinator couples.

\paragraph{Outline of the paper} 
This work begins with a review of methods for interpreting GNNs. We then describe the GNN architecture and the interpretability methods that will be employed in this study. This is followed by a simulation study performed to assess these methods' ability to identify important variables and determine the sign of their contributions to network connectivity. The simulation study is in two parts, the first involves simple simulated graphs, and the second replicates the sampling process of the Spipoll protocol. Finally, the methods are applied to the Spipoll dataset to identify which variables influence the network connectivity.

\section{Explainability for GNNs}
In the following, we provide a review of recent developments in GNNs interpretability. Since we will use a specific GNN architecture that can potentially handle sampling effects (defined in the next section) we will mainly discuss post-hoc interpretability methods. They are adapted to interpret GNNs after they have been trained, and constitute the majority of current interpretation techniques \citep{wu_interpretability_2022}. Although not the focus of this work, it is still worth mentioning some examples of interpretable models for GNNs such as graph attention networks (GAT, \citealt{velickovicGraphAttentionNetworks2018}) and disentangled representation learning for graphs \citep{ma_disentangled_2019}. Post-hoc interpretability methods can be classified into black-box interpretability methods, which do not have acces to the parameters of GNNs and their gradients, and white-box interpretability methods, which have acces to parameters and gradients. Following \cite{wu_interpretability_2022}, post-hoc interpretation methods will be presented in four categories: Approximation based methods, relevance-propagation based methods, perturbation based methods and generative explanation.

\subsection{Approximation based explanation}

The purpose of an approximation based explanation is to replace an uninterpretable GNN with an interpretable surrogate function with similar outputs. Approximation based methods can be separated into white-box and black-box methods.

\begin{sloppypar}
Among white-box methods, some are similar to methods used for vision analysis and have been adapted to graphs. \cite{baldassarre_explainability_2019} propose to estimate the squared norm of the GNN's gradient and named the method \textbf{Sensitivity Analysis}. To prevent confusion with the broader mathematical field of sensitivity analysis, which encompass a large range of techniques, we will refer to this method as \textbf{GNNSA}. Additional examples using the gradient include \textbf{GuidedBP}, \citep{baldassarre_explainability_2019}, \textbf{SmoothGrad}, \textbf{Grad$\odot$Input}, \textbf{Integrated Gradients (IG)} \citep{sanchez-lengeling2021a}, \textbf{Class Activation Mapping (CAM)} and \textbf{Grad-CAM} \citep{pope_explainability_2019}. 
\end{sloppypar}
For black-box methods, the surrogate function should be built without having access to any parameters of the GNN. Consequently, only the inputs and the corresponding outputs of the GNN are available. \textbf{GraphLime} \citep{huang_graphlime_2020} is a local explanation method for predictions on graph nodes that uses the HSIC Lasso to measure the independence between features and predictions of nodes. \textbf{RelEx} \citep{zhang_relex_2020} builds a new GNN to approximate the original GNN, before finding a minimal mask that recovers the information of the prediction. \textbf{PGM-Explainer} \citep{vu_pgm-explainer_2020} introduces an interpretable Bayesian network approximating the prediction of the GNN. \textbf{DnX} \citep{DnX} learns a surrogate GNN via knowledge distillation.

\subsection{Relevance-propagation based methods}
Relevance-propagation based approaches propagate relevance scores from high-level layers to low levels until reaching the input. Methods differ in how the score is propagated in the GNN. All these approaches are white-box methods. \textbf{Layer-wise Relevance-Propagation (LRP)}, proposed by \cite{bach_pixel_wise_2015} for image analysis, has been adapted to graphs by \cite{baldassarre_explainability_2019}. \textbf{ExcitationBP} \citep{zhang2016topdownneuralattentionexcitation} is similar to \textbf{LRP} but the relevance score is a probability distribution and the propagation is based on conditional probabilities. \textbf{GNN-LRP} identifies groups of edges that jointly contribute to prediction for the propagation. \cite{borile2023evaluatinglinkpredictionexplanations} proposes to use \textbf{Deconvolution} \citep{zeiler_visualizing_2013} to highlight which feature or edge is activated the most in the context of link predictions. \textbf{DeepLIFT}\citep{shrikumar_learning_2019} compares the activation of each neuron to its reference activation and assigns contribution scores according to the difference.

\subsection{Perturbation-based explanation}
Perturbation-based approaches assume that important features significantly influence the output, while unimportant features will not. For graph data, two types of perturbation are available. Perturbing node features by either setting them to average values or permuting them, and perturbing the graph structure by adding or removing nodes or edges. The mask modifying the graph structure can either be continuous or discrete. Most of the following methods are based on perturbing the graph structure, and all of them are black-box methods.

\textbf{GNNExplainer} \citep{ying_gnnexplainer_2019} tries to find a compact subgraph that is most crucial for prediction that maximizes mutual information between the prediction of the original graph and the prediction of the subgraph. \textbf{PGExplainer} \citep{luo_parameterized_2020} is similar to \textbf{GNNExplainer} but uses node embeddings to generate a discrete mask, while \textbf{GraphMask} \citep{schlichtkrull_interpreting_2022} also uses edge embeddings.

\textbf{CF-GNNExplainer} \citep{lucic_cf-gnnexplainer_2022} suggests building counterfactual explanations by finding the minimal number of edges to be removed such that the prediction of the GNN changes. Similarly, \textbf{CF$^2$} \citep{tan_learning_2022} uses both counterfactual and factual explanation, by also seeking a minimal set of edges/features that produce the same prediction as using the whole graph. \textbf{RCE} \citep{bajaj_robust_2022} generates robust counterfactual explanatations, where perturbation of node features should not change the estimated counterfactual subgraphs. \textbf{GCFExplainer} \citep{kosan_global_2022} estimates a small set of representative counterfactuals that globally explain all input graphs. 

\textbf{SubgraphX} \citep{yuan_SubgraphX_2021} proposes to identify important subgraphs instead of important nodes or edges. \textbf{GraphSVX} \citep{duval_graphsvx_2021} extends Shapley values \citep{Shapley1953} 
 to graphs and estimates the influence of each node and feature on the prediction.

\subsection{Generative Explanation}
Using a reinforcement learning framework, \textbf{XGNN} \citep{yuan_xgnn_2020} suggests estimating explanation by generating graphs that maximize the prediction of a given GNN model.

After reviewing the existing approaches to GNN interpretation, we present in the next session the GNN architecture employed in this study. We specify the function that requires interpretation, detail its adaptation to the Spipoll data set, and justify our selection of attribution methods. 

\section{Bipartite VGAE adaptation to the Spipoll data set}
We recall the formalism proposed in \cite{anakok2024BFGVAE} to apply variational graph auto-encoder (VGAE, \citealt{Kipf}) in the bipartite case with the specificity required for the Spipoll data set. 
\subsection{Bipartite VGAE}

\correction{A bipartite graph auto-encoder (bipartite GAE) is an embedding method composed of an encoder and a decoder. The encoder maps the two sets of nodes of a bipartite network $B$ into latent vectors $Z_1$ and $Z_2$, while the decoder reconstructs the network $\widehat{B}$ from these vectors. If $\widehat{B}$ is close to $B$, the GAE provides a meaningful embedding. In our setting, the encoder is a GNN, which updates each node's features by aggregating features from its neighbors and applying a parametric neural network, while the decoder relies on scalar products. The GAE is optimized via link prediction, where edges are removed and the GNN parameters are optimized to recover them. The variational GAE (VGAE) extends this model by enforcing a factorized Gaussian distribution in the latent space. 
}
Given \correction{a bipartite network, described by} the incidence matrix $B$ of size $n_1 \times n_2$ with covariates $X^1\in \mathbb{R}^{n_1\times d_1 }$ and $X^2\in \mathbb{R}^{n_2\times d_2 }$, let $D_1 = diag\left(\sum_{j=1}^{n_2}B_{i,j} \right),  D_2 = diag\left(\sum_{i=1}^{n_1}B_{i,j} \right)$ be respectively the row and the column degree matrices, and let $\Tilde{B} = D_1^{-\frac{1}{2}} B D_2^{-\frac{1}{2}} $ be the normalized matrix. The auto-encoder can be summarised as 

\begin{center}
\begin{tikzpicture}
    \draw node[] {$B,X_1,X_2\xrightarrow[encoder]{q(Z_1,Z_2|X_1,X_2,B)}Z_1,Z_2\xrightarrow[decoder]{p(B|Z_1,Z_2)}\widehat{B}.$} ;
\end{tikzpicture}
\end{center}

The encoder consists in associating latent variables for each node of both categories. We note by $Z_1$ a $n_1 \times D$ matrix, the rows of which $(Z_{1i}\in\mathbb{R}^D)_{1\le i\le n_1}$ are the latent variables associated to the nodes of the first category. Similarly, $Z_2$ is a $n_2 \times D$ matrix with rows  $(Z_{2j}\in\mathbb{R}^D)_{1\le j\le n_2}$ being the latent variables for nodes of the second category. the encoder is then defined as 
$$q(Z_1,Z_2|X_1,X_2,B) =  \prod_{i=1}^{n_1} q_{1}(Z_{1i}|X_1,B)\prod_{j=1}^{n_2} q_2(Z_{2j}|X_2,B) $$
where $q_1$ and $q_2$ correspond to multivariate normal distributions $\mathcal{N}(\mu,diag(\sigma^2))$.
 The parameters for the distributions $q_1$:
$(\mu_{1i},\log(\sigma_{1i}))_{1\le i\le n_1}\in \mathbb{R}^D\times \mathbb{R}^D$
are obtained by two GCN \citep{Kipf}, namely GCN$_{\mu_1}(X_1,B)$ and GCN$_{\sigma_1}(X_1,B)$
where:
$$\text{GCN}_{\mu_1}(X_1,B) = \Tilde{B}\text{ReLU}(\Tilde{B}^\top X_1 W^{(1)}_{\mu_1})W^{(2)}_{\mu_1}$$
with ReLU$(x) = max(x,0)$ and the weight matrices $W^{(k)}_{\mu_1}$ are to be estimated. GCN$_{\sigma_1}(X_1,B)$ is identically defined but with weight matrices $W^{(k)}_{\sigma_1}$. As \citep{Kipf}, we enforce that GCN$_{\mu_1}(X_1,B)$ and GCN$_{\sigma_1}(X_1,B)$ share the same first layer parameters, meaning that $W^{(1)}_{\mu_1}=W^{(1)}_{\sigma_1}$.
Symmetrically, the parameters for ${q_2: (\mu_{2j},\log(\sigma_{2j}))_{1\le j\le n_2}\in \mathbb{R}^D\times \mathbb{R}^D}$ are obtained by two GCN, namely GCN$_{\mu_2}(X_2,B)$ and GCN$_{\sigma_2}(X_2,B)$ where
$$\text{GCN}_{\mu_2}(X_2,B) = \Tilde{B}^\top\text{ReLU}(\Tilde{B} X_2 W^{(1)}_{\mu_2})W^{(2)}_{\mu_2}.$$ 
GCN$_{\sigma_2}(X_2,B)$ is identical but with weight matrices $W^{(k)}_{\sigma_2}$, and with $W^{(1)}_{\mu_2}=W^{(1)}_{\sigma_2}$.

Following \cite{rubin-delanchyStatisticalInterpretationSpectral2021}, we decide to use as a decoder the generalised random dot product
$$p(B|Z_1,Z_2) = \prod_{i=1}^{n_1}\prod_{j=1}^{n_2} p(B_{i,j}|Z_{1i},Z_{2j})$$
with 
$p(B_{i,j}|Z_{1i},Z_{2j}) = sigmoid( Z_{1i}^\top\mathbf{I}_{D_+,D_-}Z_{2j})$
where $sigmoid :x\mapsto \frac{1}{1+e^{-x}}$ and $\mathbf{I}_{D_+,D_-}$ is a diagonal matrix with $D_+$ ones followed by $D_-$ minus ones on its diagonal, such as $D_+ + D_-  = D $. The loss of the auto-encoder can be written as 
\begin{align}
     L_W  &= \mathbb{E}_{q(Z_1,Z_2|X_1,X_2,B)}[\log p(B|Z_1,Z_2)]- KL[q_1(Z_1|X_1,B)||p_1(Z_1)]\notag\\
     &-KL[q_2(Z_2|X_2,B)||p_2(Z_2)]
    \end{align}
where  $KL$ is the Kullback-Leibler divergence, and $p_1,p_2$ are Gaussian priors for $Z_1$ and $Z_2$.
From now on, this model will be referred to as \textbf{BVGAE}.

    \subsection{Connectivity prediction}
    \correction{Connectivity is the proportion of possible links between nodes that are realized.}
        Given $B$ and $X_2$, we wish to study the influence of the input $X_1$ during the learning on the expected connectivity of the output $\widehat{B}$ of the BVGAE. The expected connectivity $f_{\widehat{B}}$ can be estimated by averaging the expected probabilities of connection $\widehat{B}$ : 
        \begin{align}
            f_{\widehat{B}}(B,X_1,X_2)=f_{\widehat{B}}(X_1) &:= \frac{1}{n_1 n_2} \sum_{i=1}^{n_1}\sum_{j=1}^{n_2} \widehat{B}_{i,j}
        \end{align}

\subsection{Adaptation of BVGAE to the Spipoll data set}
\correction{The Spipoll data set provides a bipartite network $B$ where the first type of nodes is the session
of observations, and the second type corresponds to insects observed during
the session. Each session has covariates $X_1$ and a variable $P$ describing the plant genus. Link prediction task in this situation aims to predict which insect will be present during a given observation session. However, we still wish to ultimately obtain a bipartite plant-insect network, named $B'$, as this is the most widely used tool in this field of study. To construct a plant-insect
network, we propose to average all the predicted probabilities of observing an insect during a session by plants.}

For all $i$ and $j$ , $B_{i,j} \in \{0,1\}$ describes the absence or the presence of the pollinator $j$ during the session $i$. $X_1\in \mathbb{R}^{n_1\times d_1 }$ are features describing observation conditions for observation sessions. We do not consider features describing pollinators; therefore $X_2$ is set to $I_{n_2}$. Let $P = (P_{i,k}) $, $i=1,\dots,n_1,k=1,\dots,u$, where $u$ corresponds to the number of observed taxa of plants. $P_{i,k}\in \{0,1\}$ is a binarized categorical variable that describes the plant taxonomy of the $i^{th}$ session. For all $i$, there is only one coordinates $k$ such that $P_{i,k}=1$ while the others are equal to 0. To build the $u\times n_2$ binary adjacency matrix $B'$ of plant-pollinator interactions from the session-pollinator matrix $B$, we compute
$B' =\mathbf{1}( P^{\top}B>0)$.
Let $\Tilde{P}_{i,k}= \frac{P_{i,k}}{\sum_{l=1}^{n_1}P_{l,k}}$. the plant-pollinator network $B'$ can be reconstructed from $\widehat{B}$ itself, by calculating $\widehat{B'} = \Tilde{P}^\top \widehat{B}$. This is equivalent to averaging the predicted probabilities of interaction by plants. The loss function $L_W$ is adapted to also recover $B'$. 

\begin{figure}[H]
  \centering
  \includegraphics[scale=1.25]{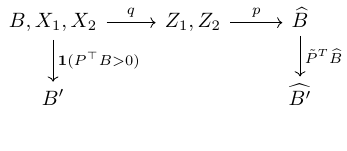}
  \caption{Summary of the model used for the training of the Spipoll data set.}
\end{figure}

As explained in \cite{anakok2024BFGVAE}, using the HSIC \cite{grettonMeasuringStatisticalDependence2005a} as a additional penalty term on the loss allows the model to construct latent variables independent of variations in variables linked to the sampling effect. Whenever this additional penalty is applied, the model will be referred to as \textbf{fair-BVGAE}.

\subsection{Connectivity prediction in the Spipoll data set}

As the main focus of this study is the plant-pollinator network connectivity, our goal is to study the feature that could impact the output $\widehat{B'}$. Given $B$ and $X_2$, we wish to study the influence of the input $X_1$ during the learning on the expected connectivity  of the output $\widehat{B'}$ of the BVGAE. The expected connectivity $f_{\widehat{B'}}$ can be estimated by averaging the expected probabilities of connection $\widehat{B'}$:
\begin{align}
f_{\widehat{B'}}(B,X_1,X_2)=f_{\widehat{B'}}(X_1) &:= \frac{1}{u n_2 } \sum_{k=1}^{u}\sum_{j=1}^{n_2} \widehat{B'}_{k,j}
\end{align}

\subsection{Choice of attribution methods for connectivity prediction}
\begin{figure}[H]
  \centering
  \includegraphics[width=\textwidth]{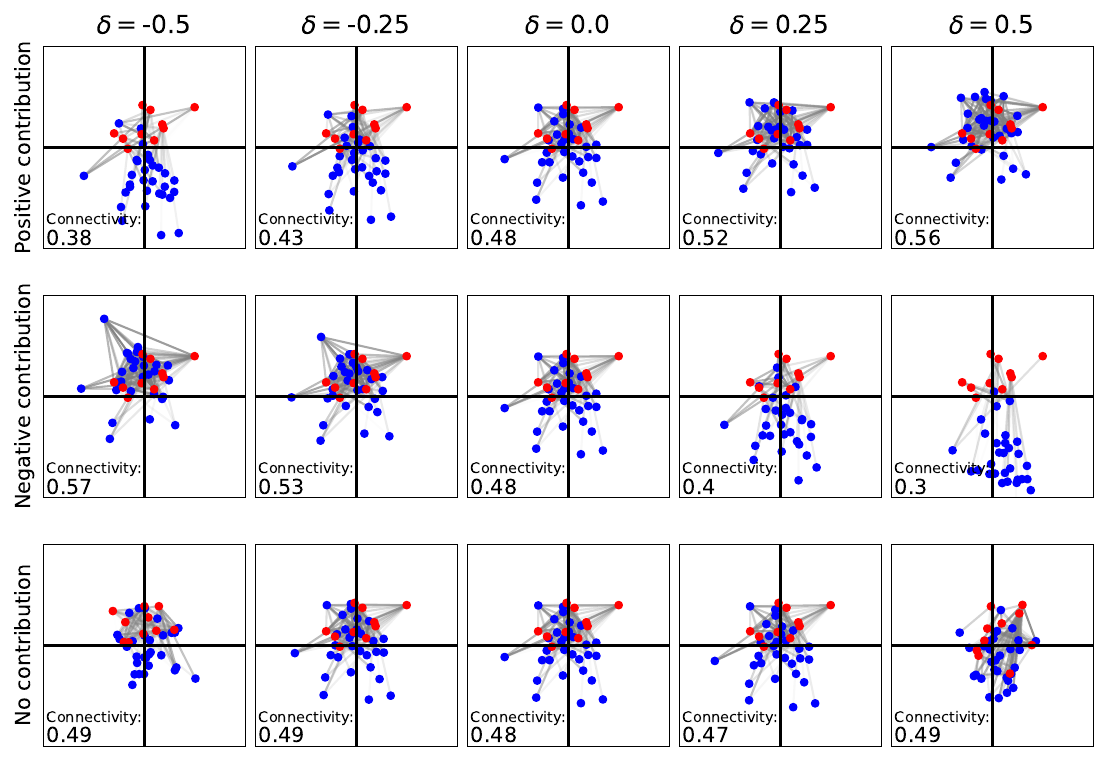}
  \caption{\correction{Insight into the behaviour we aim to detect with attribution scores. BVGAE is trained on a simulated bipartite network using features $X_1$. The resulting node embeddings are shown with different colors for the two sets of nodes. The true feature contributions to the expected connectivity $f_{\widehat{B'}}(X_1)$ are indicated on the left as "Positive contribution", "Negative contribution" and "No contribution". By perturbing each feature with a constant $\delta$, we observe how both the resulting embeddings and the expected connectivity change. The simulation settings are properly described in \cref{sec:simstu}. }}
  \label{schema_tres_simple}
\end{figure}

\correction{Attribution methods provide interpretability by assigning to each input feature a score that depends on its influence on the model’s prediction. This work is focused on
attribution methods that are able} to globally estimate a score by feature or a score for each feature specific to each node, which would be aggregated to yield a global score for each feature. Most importantly, the attribution methods should be adapted to consider continuous outputs, since connectivity is a continuous metric. \correction{An illustration of the purpose of attribution methods in this work is available in \cref{schema_tres_simple}. In this example, a BVGAE is trained on a simulated bipartite network using features $X_1$, which also provides us with an node embedding of the graph. Since the data are simulated, the true contribution of each column of $X_1$ to the expected connectivity $f_{\widehat{B'}}(X_1)$ are known. After the learning phase, for a given feature $j$, and for all nodes $i$, we offset the value of $X_{1_{i,j}}$ to $X_{1_{i,j}} + \delta$ for different values of $\delta$ and observe how both the predicted embeddings and connectivity change. For instance, for the feature corresponding to the "positive contribution", we see that increasing $\delta$ from $-0.5$ to $0.5$ also increased the expected connectivity from $0.38$ to $0.56$, meanwhile for the feature corresponding to the "negative contribution", we see that it decreased the expected connectivity from $0.57$ to $0.3$, and the "no contribution" shows only minor changes. Attribution methods should help us to detect this behaviour for more complicated cases.  }

Relevance-propagation based methods are harder to interpret, as the attributed scores are estimated on the whole GNN and not just the input layer. Most perturbation based methods, black-box approximation based methods and \textbf{XGNN} are adapted to classification tasks and give explanation to nodes or subgraphs rather than node features. \textbf{SmoothGrad}, \textbf{Grad$\odot$Input}, \textbf{IG}, \textbf{CAM}, \textbf{Grad-CAM} can estimate signed values of feature importance, while \textbf{GNNSA} can only differentiate important variables from the ones that are not, and \textbf{GuidedBP} only detects the features that positively activate the neurons. \textbf{CAM} and \textbf{Grad-CAM} require the GNN to have a specific model architecture. For these reasons, we have decided to mainly focus on the following subset of approximation based methods and one perturbation based method.

\subsubsection{Attribution methods}
In the following parts, we will adopt a simplified mathematical notation for clarity; $f$ either denotes $f_{\widehat{B}}$ or $f_{\widehat{B'}}$, and $X$ denotes $X_1$. Moreover, $X_{.,j} \in \mathbb{R}^{n_1}$ denotes the $j$-th column (or $j$-th feature) of $X$. Finally, $\Phi_j$ denotes the global contribution score for feature $j$, and $\phi_{i,j}$ denotes the contribution score of feature $j$ for node $i$.

 \textbf{SmoothGrad} is a method proposed by \cite{smilkov2017smoothgrad} to estimate saliency maps, or pixel attribution maps, to determine the importance of pixels in image recognition settings. This method straightforwardly computes the average of a noised gradient of the output of $f$ with respect to the input, and can be used to estimate a score $\phi_{i,j}$ as done by \cite{sanchez-lengeling2021a} 
$$\phi_{i,j} = \frac{1}{K} \sum_{k=1}^K \frac{\partial f(X+ E^{(k)})}{\partial x_{i,j}},\quad E^{(k)}_{i,j}\overset{i.i.d.}{\sim} \mathcal{N}(0,\sigma^2_j) $$ where $\frac{\partial f(\cdot)}{\partial x_{i,j}}$ is the gradient of $f$ along the $j$-th feature of node $i$, the noise variance $\sigma^2_j$ depends on amplitude of $X_{.,j}$. In our case, we define $\sigma_j=0.1\times(\underset{1\leq i\leq n_1}{\max}\{X_{i,j}\}-\underset{1\leq i\leq n_1}{\min}\{X_{i,j}\} )$. The method will be referred to as \textbf{Grad} throughout the remainder of this paper.

\textbf{Grad$\odot$Input} also used by \cite{sanchez-lengeling2021a}, this method gives as attribute score $\phi_{i,j}$ the element-wise product of the input node features with the gradient estimated by \textbf{Grad}. 

\textbf{Integrated Gradients} is an axiomatic attribution method proposed by \cite{sundararajanAxiomaticAttributionDeep2017}, adapted to graph by \cite{sanchez-lengeling2021a}. After defining an arbitrary baseline $X'$, the integrated gradient along the $j$-th dimension for the node $i$ for an input $X$ and baseline $X'$ is defined as 

$$\phi_{i,j} = (X_{i,j} - X'_{i,j}) \int_{0}^1 \frac{\partial f(\alpha X + (1-\alpha) X')}{\partial x_{i,j}} d\alpha.  $$

\textbf{GraphSVX} Graph Shapley Explanations for GNN has been developed by \cite{duval_graphsvx_2021}, who adapted the Shapley value theory for graphs. This model is able to attribute a feature importance for each node and each node feature, by first constructing a data set $\mathcal{D}:= \{(z,f(z')\}$ where $z = (z^{n_1},z^{d_1})$ is a mask representing the selected nodes and node features, and $z'$ is the subgraph of only selected node, where unselected node features are set to average value. Once $\mathcal{D}$ is constructed, a weighted linear regression, which is the explicable surrogate, 
 is adjusted and the estimated coefficients $\Phi_j$ correspond to feature importance. In our setting, we discard the importance of the nodes and only consider the feature importance of covariates.

\subsubsection{Aggregate node score}
\textbf{Grad}, \textbf{Grad$\odot$Input} and \textbf{Integrated Gradients} give attribution score $\phi_{i,j}$ for the $j$-th feature of node $i$. 
For a given feature $j$ we estimate a global feature score $\Phi_{j}$ by taking the average value $\Phi_{j}= \frac{1}{n_1}\sum_{i=1}^{n_1}\phi_{i,j}$.

If the nodes are partitioned into $K$ groups (e.g. by taxa of plants for the Spipoll data set), the score can also be aggregated by group by averaging attribution score on each group of nodes. The result $\Phi_{k,j}$ would be represented by a $K\times d_1$ matrix. However, \textbf{GraphSVX} is not adapted for such estimation. We propose to slightly change \textbf{GraphSVX} to overcome this issue, by first constructing a data set $\mathcal{D}:= \{(z,f(z'_k)\}$ where $z = (z^{n_1},z^{d_1})$ is a mask representing the selected nodes and node features, and $z'_k$ is the subgraph of only selected node, where unselected node features, and nodes outside of group $k$, are set to average value. The estimated coefficient $\Phi_{k,j}$ corresponds to feature importance $j$ for each group $k$.

\section{Simulation study}
\label{sec:simstu}

This simulation study is divided into two major parts. The first part is performed on simulated bipartite networks. The second part aims to numerically mimick the spipoll sampling \correction{process}. 
 Both parts consist of a series of various simulations with progressively increasing complexity. After describing the simulation settings, results are presented with tables and graphics. While this work presents only a subset of the simulations, the complete simulation set is accessible on GitHub : \url{https://github.com/AnakokEmre/graph_features_importance}.

\subsection{Evaluation metrics}
This study focuses on two factors in the simulation. The sign of the score should indicate whether the feature contributes positively or negatively to the expected connectivity, and its magnitude should distinguish features that truly contribute to the expected connectivity from those that do not. \correction{For example, in the simulation performed in \cref{schema_tres_simple}, the score associated with the feature in the first row is expected to be positive, whereas the score for the feature in the middle row should be negative. The score of the variable in bottom row is expected to be smaller in magnitude, remaining closer to zero.}

\correction{Given the increasing complexity of our simulation, this work mainly focuses on the following quantitative metrics :}
\begin{itemize}
    \item $+$ represents the proportion of features with positive contributions that have been correctly identified as positive.
    \item $-$ represents the proportion of features with negative contributions that have been correctly identified as negative.
    \item $AUC$ denotes the area under the ROC, which is calculated with the absolute values of the estimated scores and the ground truth. 
\end{itemize}

\subsection{Settings on simulated bipartite networks}\label{settings_on_simulated_bipartite_networks}

In the following simulations, we generate bipartite networks by first simulating the corresponding latent space. The latent space, or a transformation of it, will be used as a covariate in the model. The key difference between simulation settings is not the network generation method, but the manner in which the available covariates are incorporated into the model.

Let $n_1= 1000$ and $n_2=100$, let $D_+$ be an integer, let $D_- = D_+$ and let $D=D_+ + D_-$. Let $Z_1^+\in\mathbb{R}^{n_1 \times D_+}$ and $Z_1^-\in\mathbb{R}^{n_1 \times D_-}$ such as $Z_{1_{i,j}}^+\overset{i.i.d.}{\sim} \mathcal{N}(0,1)$ and  $Z_{1_{i,j}}^-\overset{i.i.d.}{\sim} \mathcal{N}(0,1)$ independent of $Z_1^+$. Let $Z_1 = \left[Z_1^+| Z_1^- \right]$ be the concatenation of  $Z_1^+$ and  $Z_1^-$. Let $Z_2\in\mathbb{R}^{n_2 \times D}$ such as $Z_{2_{i,j}}^+\overset{i.i.d.}{\sim} \mathcal{N}(1,1)$. For $1\leq i\leq n_1$, $Z_{1i}\in\mathbb{R}^{D} $ represents the $i$-th row of $Z_1$. Similarly, $Z_{2j}\in\mathbb{R}^{D} $ represents the $j$-th row of $Z_2$. Finally, our bipartite adjacency matrix is simulated with a Bernoulli distribution $ B_{i,j} \overset{i.i.d.}{\sim} \mathcal{B}(sigmoid(Z_{1i}^\top\mathbf{I}_{D_+,D_-}Z_{2j}))$. $Z_1$ and $Z_2$ are, respectively, row nodes and column nodes latent representations of the generated network. Given how the network is constructed, higher values of $Z_1^+$ are expected to be positively correlated with connectivity, while higher values of $Z_1^-$ are expected to be negatively correlated with connectivity. 

\textbf{Inputs :} In the following part, we summarize how we construct the observed $X_1$ that will be used as a covariate to fit the model. To improve readability, we will change the notation from $Z_1$ to $Z$. Let $D_0$ be an integer and let $X_1^0 \in \mathbb{R}^{n_1 \times D_0}$ be a noise matrix  with ${X}^0_{i,j}\overset{i.i.d.}{\sim} \mathcal{N}(0,1)$. The $n_1$ row nodes are partitioned into $K$ groups, node $i$ belongs to group $Q[i]$ with $Q[i]\in \{1,\dots,K\}$. If there are no groups, then $K=1$. For $1\leq k \leq K, 1\leq j \leq D$, $\gamma_{k,j}\in \{-1,0,1\}$ describes the combined effect of group $k$ on the covariate $j$. The set of value taken by $\gamma$ can change depending on the simulation setting. For $1\leq i \leq n_1, 1\leq j \leq D$, let $X$ such as
$$
 X_{i,j}= \left\{
    \begin{array}{ll}
        \gamma_{Q[i],j}Z_{i,j}& \mbox{if } \gamma_{Q[i],j}\neq 0 \\
        \\
        \xi_{i,j}\mbox{ with } \xi_{i,j}\overset{i.i.d.}{\sim} \mathcal{N}(0,1) & \mbox{if }  \gamma_{Q[i],j}=0
    \end{array}
\right.
$$
Finally, we define $$X_1 = \left[H|X|X^0 \right]$$ where $H$ is either $\mathbf{1}_{n_1}$ or $\left[\mathbf{1}_{n_1}| P \right]$. 
We set $X_2 = \mathbf{1}_{n_2}$.  BVGAE is trained with adjacency matrix $B$ and covariates $X_1,X_2$, which means that there are two sources of noise, noises coming from $X_1^0$, and another from features where $\gamma_{Q[i],j}=0$. The learning can also be done with the fair-BVGAE. In this case, some input columns of $X$ are selected and the learning is penalized by the HSIC between the estimated latent space and these columns. Once the model is trained, previously described attribution methods are fit on $f_{\widehat{B}}(X_1)$ to study the impact of the features of $X_1$ on connectivity.

All the available simulation settings in the supplementary are displayed in \cref{setting_table1}. In this paper, only the results for settings \textbf{(1.A)},\textbf{(1.B)},\textbf{(1.C)} and \textbf{(1.D)} are presented.
\begin{table}[h!]

    \centering
    \rowcolors{2}{gray!20}{white}
    \begin{tabu}to \linewidth {>{}X|>{\centering}X>{\centering}X>{\centering}X>{\centering}X>{\centering}X>{\centering}X}
\toprule
Settings & $D_+$ & $D_0$ &$K$ & $\Gamma$ & HSIC& $H$\\
\midrule
\correction{\cref{schema_tres_simple}} & \correction{2} & \correction{2}  & \correction{1} & \correction{$\{1\}$}       & \correction{\_} & \correction{$\mathbf{1}_{n_1}$} \\
0 & 3 & 3  & 1 & $\{1\}$       & \_ & $\mathbf{1}_{n_1}$ \\
1 & 3 & 3  & 1 & $\{1\}$       & \_ & $\mathbf{1}_{n_1}$ \\
2 \textbf{(1.A)} & \textbf{3} & \textbf{50} & \textbf{1} & $\mathbf{\{1\}}$       & \_ & $\mathbf{1}_{n_1}$ \\
3 & 3 & 50 & 1 & $\{1\}$       & \_ & $\mathbf{1}_{n_1}$ \\
4 \textbf{(1.B)}& \textbf{1} & \textbf{1}  & \textbf{2} & $\mathbf{\{1,-1\}}$    & \_ & $\mathbf{1}_{n_1}$ \\
5 & 3 & 50 & 2 & $\{1,-1\}$    & \_ & $\mathbf{1}_{n_1}$ \\
6 & 3 & 6  & 2 & $\{1,0,-1\}$  & \_ & $\mathbf{1}_{n_1}$ \\
7 & 3 & 6  & 2 & $\{1,-1\}$    & \_ & $[\mathbf{1}_{n_1},P]$ \\
8 \textbf{(1.C)}& \textbf{3} & \textbf{6}  & \textbf{2} & $\mathbf{\{1,0,-1\}}$  & \_ & $\mathbf{1}_{n_1}$ \\
9 & 3 & 6  & 4 & $\{1,0,-1\}$  & \_ & $[\mathbf{1}_{n_1},P]$ \\
10\textbf{(1.D)}& \textbf{3} & \textbf{1}  & \textbf{4} & $\mathbf{\{1\}}$      & \textbf{2}  & $\mathbf{1}_{n_1}$ \\
11 & 4 & 50 & 2 & $\{1,-1\}$   & 2  & $\mathbf{1}_{n_1}$ \\
12 & 4 & 8  & 4 & $\{1,0,-1\}$ & 2  & $\mathbf{1}_{n_1}$ \\
13 & 4 & 8  & 4 & $\{1,0,-1\}$ & 2  & $[\mathbf{1}_{n_1},P]$ \\
\bottomrule
\end{tabu}
\caption{Parameters for the presented simulation settings. We remind that the true latent space is of size $D = D_+ + D_-$ with $D_+ = D_- $, and $D_0$ is the number of the noise covariates. $K$ is the number of groups. $\Gamma$ represents the set of values possibly taken by $\gamma$. The HSIC columns determine how many columns of $X$ are penalized by the HSIC during the learning of the fair-BVGAE, empty values correspond to classical BVGAE learning. $H$ corresponds to an additional covariate used for the learning.}\label{setting_table1}
\end{table}

Simulation 1.A correspond to a simple simulation setting, where there is only one effect by covariates, but with a lot of noise. Simulation 1.B correspond to a simulation where a covariates can have two effects depending on the group of nodes. Simulation 1.C corresponds to a harder simulation with additional covariates effect and a larger number of groups. Simulation 1.D is similar to 1.A but also uses the HSIC loss to study its impact on the attributed score.

\textbf{Expectation :} Attribution scores will be aggregated according to the groups given by $Q$. The scores should be either negative or positive depending on the values of $\gamma$. Attribution scores estimated for $X$  where $\gamma \neq 0$ should have higher magnitude to those estimated for $X^0$, those where $\gamma=0$, or columns penalized by the HSIC. The value associated with $\mathbf{1}_{n_1}$ is not taken into account for the evaluation metrics.

\subsection{Results}

\begin{table}
    \centering\scalebox{1}{
    \begin{tabu} to \linewidth {>{\raggedright}X>{\raggedleft}X>{\raggedleft}X>{\raggedleft}X>{\raggedleft}X}
        \hline
        \multicolumn{1}{c}{} & \multicolumn{1}{c}{GraphSVX} & \multicolumn{1}{c}{Grad} & \multicolumn{1}{c}{Grad $\odot$ Input} & \multicolumn{1}{c}{IG}\\
        \hline
        \multicolumn{5}{l}{\textbf{1.A}}\\
        \hline
        \multicolumn{1}{c}{\hspace{1em}+} & \multicolumn{1}{c}{\textbf{1.000}} & \multicolumn{1}{c}{\textbf{1.000}} & \multicolumn{1}{c}{\textbf{1.000}} & \multicolumn{1}{c}{\textbf{1.000}}\\
        \hline
        \multicolumn{1}{c}{\hspace{1em}-} & \multicolumn{1}{c}{0.000} & \multicolumn{1}{c}{\textbf{1.000}} & \multicolumn{1}{c}{0.011} & \multicolumn{1}{c}{0.000}\\
        \hline
        \multicolumn{1}{c}{\hspace{1em}AUC} & \multicolumn{1}{c}{0.997} & \multicolumn{1}{c}{0.879} & \multicolumn{1}{c}{0.976} & \multicolumn{1}{c}{\textbf{1.000}}\\
        \hline\addlinespace[0.5cm]
        \multicolumn{5}{l}{\textbf{1.B}}\\
        \hline
        \multicolumn{1}{c}{\hspace{1em}+} & \multicolumn{1}{c}{\textbf{0.600}} & \multicolumn{1}{c}{0.467} & \multicolumn{1}{c}{0.467} & \multicolumn{1}{c}{0.533}\\
        \hline
        \multicolumn{1}{c}{\hspace{1em}-} & \multicolumn{1}{c}{0.533} & \multicolumn{1}{c}{\textbf{0.844}} & \multicolumn{1}{c}{0.544} & \multicolumn{1}{c}{0.433}\\
        \hline
        \multicolumn{1}{c}{\hspace{1em}AUC} & \multicolumn{1}{c}{0.546} & \multicolumn{1}{c}{0.267} & \multicolumn{1}{c}{0.608} & \multicolumn{1}{c}{\textbf{0.733}}\\
        \hline\addlinespace[0.5cm]
        \multicolumn{5}{l}{\textbf{1.C}}\\
        \hline
        \multicolumn{1}{c}{\hspace{1em}+} & \multicolumn{1}{c}{0.500} & \multicolumn{1}{c}{\textbf{0.867}}& \multicolumn{1}{c}{0.527} & \multicolumn{1}{c}{0.667}\\
        \hline
        \multicolumn{1}{c}{\hspace{1em}-} & \multicolumn{1}{c}{0.497} & \multicolumn{1}{c}{0.527} & \multicolumn{1}{c}{\textbf{0.533}} & \multicolumn{1}{c}{0.387}\\
        \hline
        \multicolumn{1}{c}{\hspace{1em}AUC} & \multicolumn{1}{c}{0.508} & \multicolumn{1}{c}{0.326} & \multicolumn{1}{c}{0.569} & \multicolumn{1}{c}{\textbf{0.572}}\\
        \hline\addlinespace[0.5cm]
        \multicolumn{5}{l}{\textbf{1.D}}\\
        \hline
        \multicolumn{1}{c}{\hspace{1em}+} & \multicolumn{1}{c}{0.983} & \multicolumn{1}{c}{\textbf{1.000}} & \multicolumn{1}{c}{0.850} & \multicolumn{1}{c}{0.967}\\
        \hline
        \multicolumn{1}{c}{\hspace{1em}-} & \multicolumn{1}{c}{0.017} & \multicolumn{1}{c}{\textbf{1.000}} & \multicolumn{1}{c}{0.083} & \multicolumn{1}{c}{0.017}\\
        \hline
        \multicolumn{1}{c}{\hspace{1em}AUC} & \multicolumn{1}{c}{\textbf{0.942}} & \multicolumn{1}{c}{0.853} & \multicolumn{1}{c}{0.803} & \multicolumn{1}{c}{0.939}\\
        \hline
    \end{tabu}}
    \caption{Average results on 30 simulated bipartite networks. "$+$" (resp. "$-$") represents the proportion of features with positive (resp. negative) contributions that have been correctly identified as positive (resp. negative). The best score in each row is indicated in bold.}\label{table_simuFI1}
\end{table}

\begin{figure}[H]
    \centering
    \includegraphics[width =0.7 \textwidth]{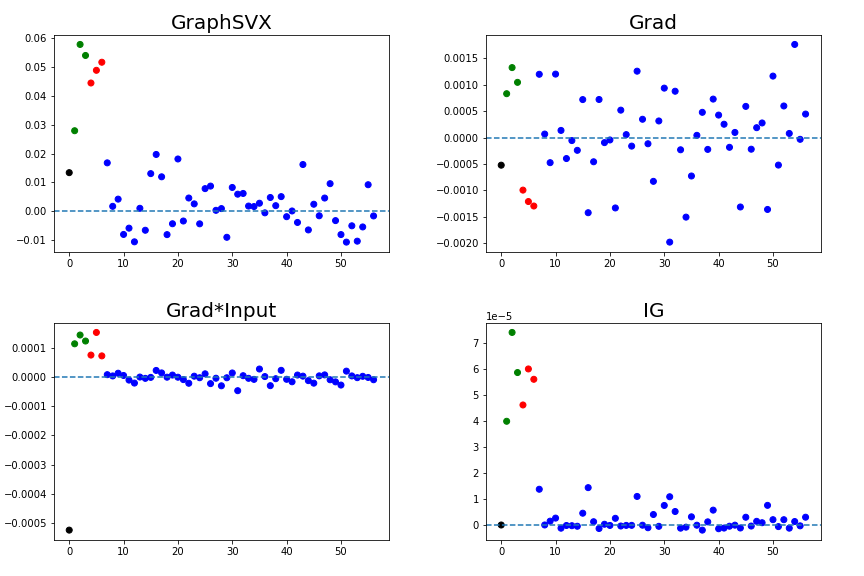}
    \caption{Estimated feature importance in Simulation 1.A. for a single run. The dashed line is positioned at zero. The black dot represents the estimated score for $\mathbf{1}_{n_1}$. The green (resp. red) dots represent the estimated score for features where positive (resp. negative) values were expected. The blue dots are scores attributed to noise.
    }
    \label{plot_simuFI1}
\end{figure}

\begin{figure}[H]
    \centering
    \includegraphics[width = 0.7\textwidth]{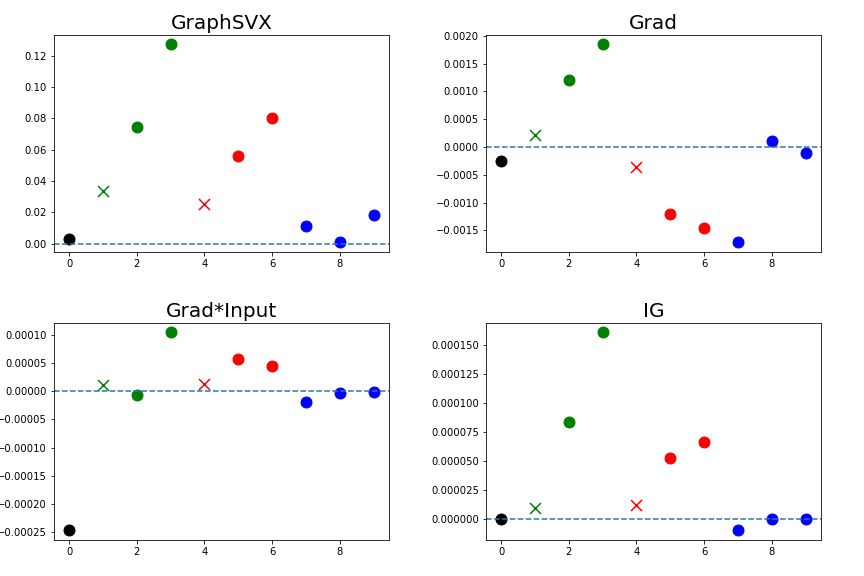}
    \caption{Estimated feature importance in Simulation 1.D for a single run. The dashed line is positioned at zero. The black dot represents the estimated score for $\mathbf{1}_{n_1}$. The green (resp. red) dots represent the estimated score for features where positive (resp. negative) values were expected. The blue dots are scores attributed to noise. Variables penalized by the HSIC are represented with a cross.}
    \label{plot_simuFI3}
\end{figure}

\begin{figure}
    \centering
    \includegraphics[width = 0.8\textwidth]{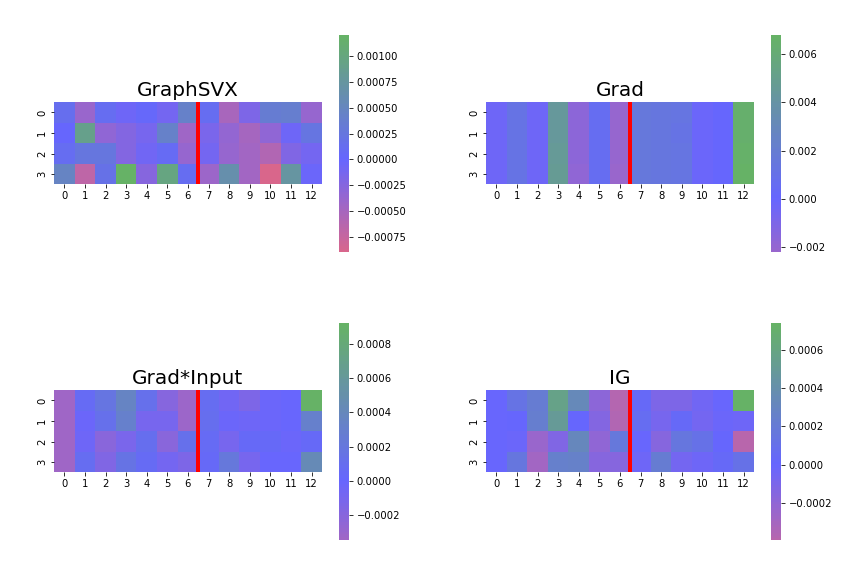}
    \includegraphics[width = 0.8\textwidth]{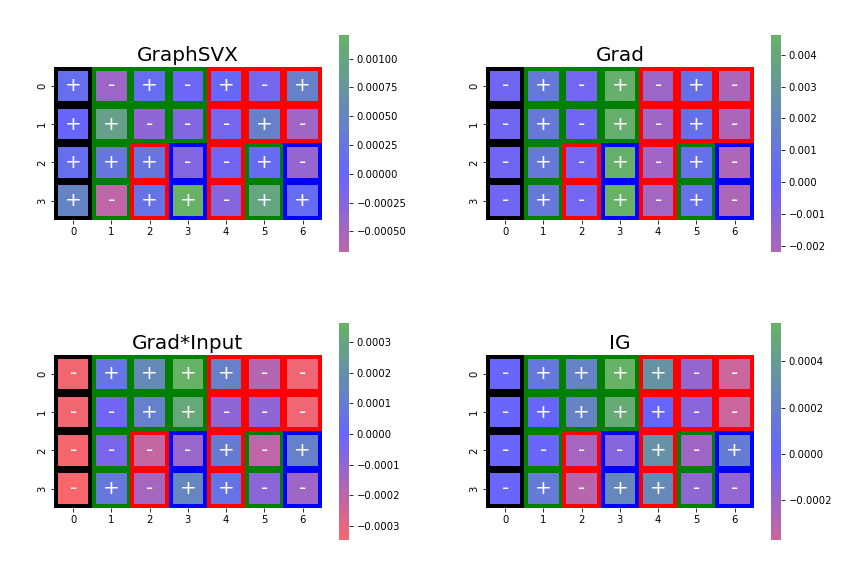}
    \caption{Estimated feature importance in Simulation 1.C for a single run. For all plot, each row represents a group, while each column represents a feature. The top six graphics display the estimated scores for all features. Features to the right of the red lines are noise. The bottom six graphics are zoomed-in sections of the left portion of the top six graphics. For each cell, the border frame represents the expected value, while the interior represents the estimated value. The black frames represent the estimated score for $\mathbf{1}_{n_1}$, green (resp. red) frames represent the score for features where positive (resp. negative) values were expected, and the blue frames are scores attributed to noise. The sign "+" or "-" denotes the sign of the estimated score within each cell.}
    \label{plot_simuFI2}
\end{figure}

As we can see in \cref{table_simuFI1}, results differ depending on the simulation settings and the used methods. For simulation 1.A (\cref{plot_simuFI1}), only \textbf{Grad} retrieves correctly the sign of the effect, but its AUC is slightly lower than the ones from the other methods. 
If we give a closer look at \textbf{GraphSVX}, \cref{table_simuFI1} 1.A reveals that the method detects correctly important features with an AUC close to 1. In \cref{plot_simuFI1}, where the blue points represent noise, it assigns more importance in absolute value to features that truly contribute to connectivity, which are indicated with green and red points. However, although we expected negative values for the red points, \textbf{GraphSVX} assigns a positive sign to all important variables, including those that have a negative effect on connectivity, as we can see by "$-$" being equal to 0 in \cref{table_simuFI1} (1.A). In setting 1.A, all methods manage to distinguish the variables that affect the connectivity from the ones that are noise. However, in simulations 1.B and 1.C (\cref{plot_simuFI2}), where covariates can have multiple effects depending on the group of the node, no method can consistently estimate the sign of the effect. On average, \textbf{Grad} has better sign accuracy, while \textbf{Grad$\odot$Input} and \textbf{IG} can better detect significant variable from the noise. The proposed aggregation method for \textbf{GraphSVX} does not seem to detect correctly the combined effect of the group and the covariate. In simulation 1.D, the scores are correctly adjusted to the expected behavior, the HSIC-penalized variables have in average lower scores than the others, which is the expected result. This is illustred in (\cref{plot_simuFI3}).

To conclude this first simulation set, combining the results of different attribution methods can allow us to retrieve the correct sign and the correct magnitude of covariates if they only have one effect on the outcome. However, if the covariates have multiple effect depending on the node group, it's much harder to evaluate which covariates have more effect than the others, and it is not possible to correctly determine the sign.

\subsection{Settings on simulated sampling process}
\label{setsimulatedsamplingprocess}

 This simulation study tries to replicate numerically the sampling process taking place in the Spipoll data set. Covariates will be used to explain the observation probabilities. The key difference between simulation settings is not the network generation method, but the manner in which the available covariates are incorporated into the model.

 \textbf{Underlying plant-pollinator network : } An underlying plant-insect network $B_0'$ is generated in order to account for possible interactions. It consists of a bipartite SBM made of $u=83$ plants and $n_2= 306$ insects, with parameters $$\alpha =  (0.3,0.4,0.3),\quad \beta= (0.2,0.4,0.4),\quad  \pi = \begin{bmatrix} 0.95 & 0.80 & 0.50\\ 0.90 & 0.55 & 0.20\\  0.70 & 0.25 & 0.06 \end{bmatrix}, $$ where $\alpha$ is the row groups proportion, $\beta$ the columns group proportion $\pi$ denote the connectivity matrix.
 This means that for each plant $k$ (resp. insect $j$), there is a latent variable $V^1_k \in \{1,2,3\}$ (resp. $V^2_j \in \{1,2,3\}$) such as  $V^1_k$ follows a multinomial distribution $Mult(1,\alpha)$, $V^2_j \sim Mult(1,\beta)$ and the probability of having an interaction between plant $k$ and insect $j$ is given by $\mathbb{P}(B_{0 k,j}'=1 |V^1_k,V^2_j) = \pi_{V^1_k,V^2_j}$. The given parameters  correspond to a nested network, a model often encountered in ecological studies.

\begin{figure}
  \centering
  \includegraphics[width = 0.95\textwidth]{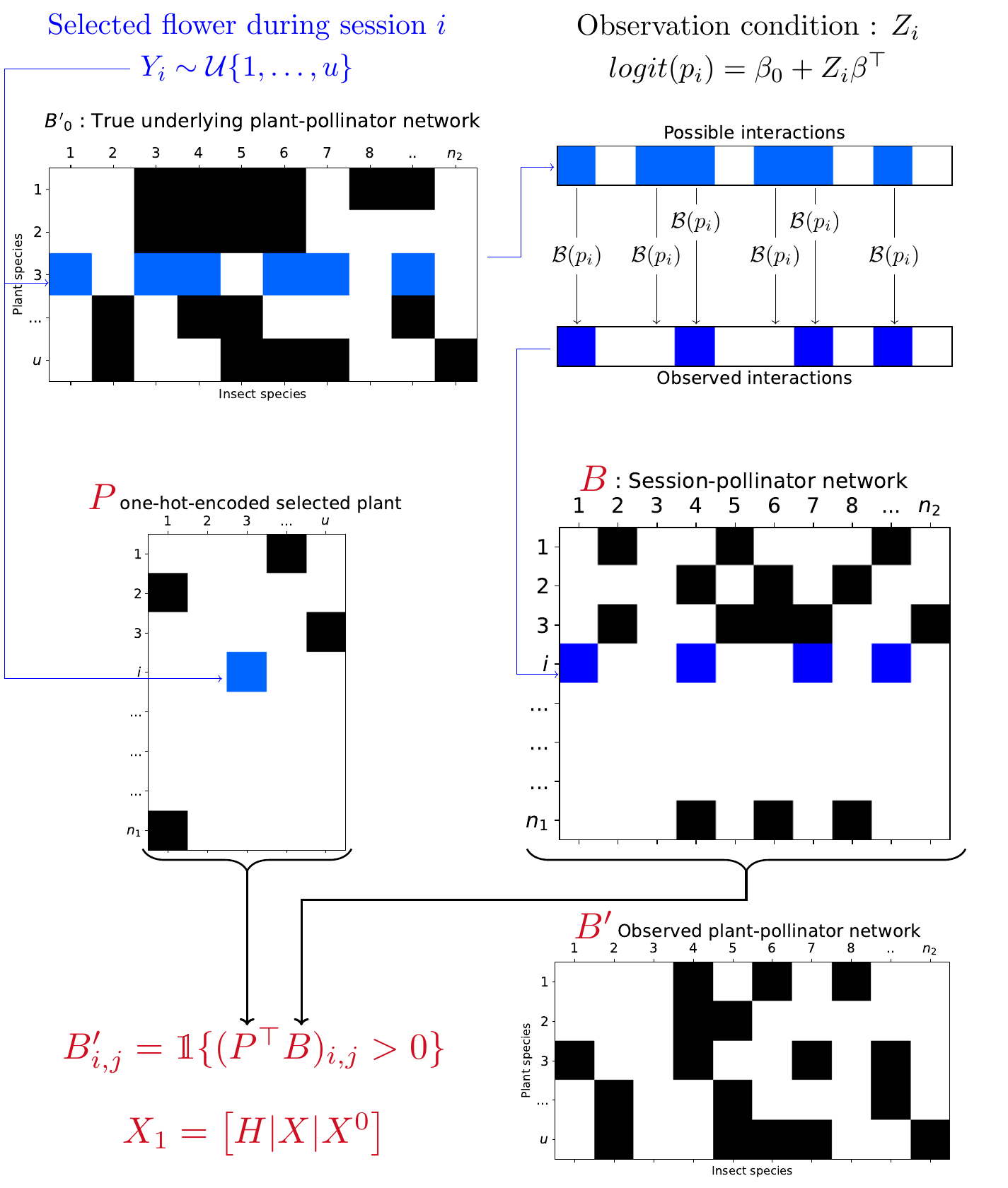}
  \caption{Numerical replication of the Spipoll sampling process. In session $i$, the user randomly selects a plant $Y_i$ from a uniform distribution. The visible interactions are determined by the matrix $B'_0$; in this example, the user has selected the plant corresponding to the blue row. Given observation condition $Z_i$, the probability of observing each possible interaction is $p_i$. The observed interactions are stored in the session-pollinator network $B$. Knowing the selected plant, we can construct the matrix $P$, which represents the plant chosen in each session. From $B$ and $P$ the observed plant-pollinator network $B'$ can be deduced. Only variables highlighted in red are considered observed and can be used for training the model. Construction of $X_1$ is detailed in the \textbf{Inputs} section.}\label{sampling_process_FI}
  \end{figure}

 \textbf{Session-pollinator network : } Let $n_1 =1000$ be the number of observers. Each user will select uniformly at random one plant species $Y_i$, and will observe possible interactions from the $Y_i$-th row of the matrix $B_0'$ at random, with a probability defined as followed : Let $Z^+\in\mathbb{R}^{n_1 \times D_+}$ and $Z^-\in\mathbb{R}^{n_1 \times D_-}$ such as $Z_{i,j}^+\overset{i.i.d.}{\sim} \mathcal{N}(0,1)$ and  $Z_{i,j}^-\overset{i.i.d.}{\sim} \mathcal{N}(0,1)$ independent of $Z^+$. Let $Z = \left[Z^+| Z^- \right]$ be the concatenation of $Z^+$ and $Z^-$. Let $\beta = (1,\dots,1, -1,\dots,-1)$ a vector whose $D_+$ first coordinates are $1$ and $D_-$ next coordinates are $-1$, and let $\beta_0 \in \mathbb{R}$. Let $p$ be the $n_1$ sized vector such as 
 $$logit(p) = \beta_0\mathbf{1}_{n_1} +  Z\beta^\top,$$ where $logit(p)= \begin{pmatrix}
    logit(p_1) \\
    \vdots \\
    logit(p_{n_1})
  \end{pmatrix}$ is the vector made of element wise application of the logit function. Finally, the probability that user $i$ sees insect $j$ in front of flower $Y_i$ is given by the network of possible interactions $B_{0}'$ and the probability of observation $p$ that depends on condition of observation $Z$, with 
  \begin{align}
    Y_i \overset{i.i.d.}{\sim}& \mathcal{U}\{1,\dots,u\}\\
    B_{i,j}| {Y_i=k} \overset{i.i.d.}{\sim}& \mathcal{B}(p_i B_{0_{k,j}}'),
\end{align}
where ${U}\{1,\dots,u\}$ is the discrete uniform distribution on $\{1,\dots,u\}$. The user can not see insect $j$ on flower $k$ if  $B_{0_{k,j}}'=0$, and otherwise the insect can be observed with probability $p_i$.
 Once the observations-insects network $B$ is constructed, we also have access to $P_{i,k}$, the one-hot-encoded categorical variable that describes the plant taxonomy of the $i$-th sessions, and consequently to the observed plant-pollinator network $B'$. A summary of the procedure is available in \cref{sampling_process_FI}.

  Given how the network is constructed, higher values of $Z^+$ are expected to be positively correlated with connectivity, while higher values of $Z^-$ are expected to be negatively correlated with connectivity. The following simulations settings are similar to the previous ones, but are adapted to the Spipoll simulation. 
 
  \textbf{Inputs :} The inputs are identically created as in \cref{settings_on_simulated_bipartite_networks}. The $n_1$ sessions are partitioned into $K$ groups, session $i$ belongs to group $Q[i]$ with $Q[i]\in \{1,\dots,K\}$, $\gamma_{k,j}$ describes the combined effect of group $k$ on the covariate $j$ and 
  \begin{align}
 \label{eq_X_1_gamma}
  X_{i,j}= \left\{
     \begin{array}{ll}
         \gamma_{Q[i],j}Z_{i,j}& \mbox{if } \gamma_{Q[i],j}\neq 0 \\
         \\
         \xi_{i,j}\mbox{ with } \xi_{i,j}\overset{i.i.d.}{\sim} \mathcal{N}(0,1) & \mbox{if }  \gamma_{Q[i],j}=0
     \end{array}
 \right.
 \end{align}
  Finally, we define 
  \begin{align}
     \label{eq_X_1_FI}
     X_1 = \left[H|X|X^0 \right]
  \end{align} where $H$ is either $\mathbf{1}_{n_1}$ or $\left[\mathbf{1}_{n_1}| P \right]$. Even if it will be the latest assumption, the number of groups $K$ is not necessarily tied to the number of plants $u$. With $X_2 = \mathbf{1}_{n_2}$, BVGAE adaptation to the Spipoll data set is trained with adjacency matrix $B$, plant matrix $P$
  and covariates $X_1,X_2$. The learning can also be done with the fair-BVGAE. In this case, some input columns of $X_1$ are selected and the learning is penalized by the HSIC between the estimated latent space and these columns. Once the model trained, previously described attribution methods are fit on $f_{\widehat{B'}}(X_1)$ to study the impact of the features of $X_1$ on connectivity. 
 
  All the available simulation settings in the supplementary are displayed in \cref{setting_table1}. In this paper, only the results for settings \textbf{(2.A-E)} are presented.
 
  \begin{table}
     \centering
     \begin{tabu}to \linewidth {>{}X|>{\centering}X>{\centering}X>{\centering}X>{\centering}X>{\centering}X>{\centering}X}
 \toprule
 Settings & $D_+$ & $D_0$ &$K$ & $\Gamma$ & HSIC & $H$ \\
 \midrule
 0  & 3 & 3  & 1  & $\{1\}$       & \_ & $\mathbf{1}_{n_1}$ \\
     1  & 3 & 3  & 1  & $\{1\}$       & \_ & $\mathbf{1}_{n_1}$ \\
     \rowcolor{gray!20}
     2 \textbf{(2.A)} & \textbf{3} & \textbf{50} & \textbf{1}  & $\mathbf{\{1\}}$       & \_ & $\mathbf{1}_{n_1}$ \\
     3  & 3 & 50 & 1  & $\{1\}$       & \_ & $\mathbf{1}_{n_1}$ \\
     4  \textbf{(2.B)} & \textbf{1} & \textbf{1}  & \textbf{2}  & $\mathbf{\{1,-1\}}$    & \_ & $\mathbf{1}_{n_1}$ \\
     5  & 3 & 50 & 2  & $\{1,-1\}$    & \_ & $\mathbf{1}_{n_1}$ \\
     6  & 3 & 6  & 2  & $\{1,0,-1\}$  & \_ & $\mathbf{1}_{n_1}$ \\
     7  & 3 & 6  & 2  & $\{1,-1\}$    & \_ & $[\mathbf{1}_{n_1},P]$ \\
     8  \textbf{(2.C)} & \textbf{3} & \textbf{6}  & \textbf{2}  & $\mathbf{\{1,0,-1\}}$  & \_ & $\mathbf{1}_{n_1}$ \\
     9  & 3 & 6  & 4  & $\{1,0,-1\}$  & \_ & $[\mathbf{1}_{n_1},P]$ \\
     10\textbf{(2.D)}& \textbf{3} & \textbf{1}  & \textbf{4}  & $\mathbf{\{1\}}$       & \textbf{2}  & $\mathbf{1}_{n_1}$ \\
     11 & 4 & 50 & 2  & $\{1,-1\}$    & 2  & $\mathbf{1}_{n_1}$ \\
     12 & 4 & 8  & 4  & $\{1,0,-1\}$  & 2  & $\mathbf{1}_{n_1}$ \\
     13 & 4 & 8  & 4  & $\{1,0,-1\}$  & 2  & $[\mathbf{1}_{n_1},P]$ \\
     14\textbf{(2.E)}& \textbf{4} & \textbf{8}  & \textbf{83} & $\mathbf{\{1,0,-1\}}$  & \textbf{2}  & $[\mathbf{1}_{n_1},P]$ \\
     15\textbf{(2.F)}& \textbf{4} & \textbf{50} & \textbf{83} & $\mathbf{\{1,0,-1\}}$  & \textbf{2}  & $[\mathbf{1}_{n_1},P]$ \\
 \bottomrule
 \end{tabu}
 \caption{Parameters for the presented simulation settings. We remind that the number of parameters to generate the model is $D = D_+ + D_-$ with $D_+ = D_- $, and $D_0$ is the number of the noise covariates. $K$ is the number of groups. $\Gamma$ represents the set of value possibily taken by $\gamma$. The HSIC columns determine how much columns of $X_1$ are penalized by the HSIC during the learning of the fair-BVGAE, empty values correspond to classical BVGAE learning. $H$ corresponds to an additional covariate used for the learning.}\label{setting_table2}
 \end{table}

 Simulations 2.A-2.D are the Spipoll adaptation of simulations 1.A-1.D. Simulations 2.E-2.F combine the settings of simulation 2.C and 2.D with additional variables, more noise, and a greater number of groups. These groups correspond to those formed by the plant matrix $P$. These settings are the most complicated but they are the closest to represent the data available at hand for the Spipoll data set. Simulation 2.E and 2.F only differ by the size of the noise matrix.
 \newpage
 \textbf{Expectation :} Attribution scores will be aggregated according to the groups given by $Q$. The scores should be either negative or positive depending on the values of $\gamma$. Attribution scores estimated for $X$  where $\gamma \neq 0$ should have higher magnitude to those estimated for $X^0$, those where $\gamma=0$, or columns penalized by the HSIC. The values associated with $H$ are not taken into account for the evaluation metrics.
 
 \subsection{Results}
 
 \setlength{\tabcolsep}{12pt}
 \begin{table}[H]
     \centering
     \scalebox{0.8}{
 \begin{tabu} to \linewidth {>{\raggedright}X>{\raggedleft}X>{\raggedleft}X>{\raggedleft}X>{\raggedleft}X}
 \hline
 \multicolumn{1}{c}{ } & \multicolumn{1}{c}{GraphSVX } & \multicolumn{1}{c}{Grad} & \multicolumn{1}{c}{Grad x Input} & \multicolumn{1}{c}{IG} \\
 \hline
 \multicolumn{5}{l}{\textbf{2.A}}\\
 \hline
 \multicolumn{1}{c}{\hspace{1em}+} & \multicolumn{1}{c}{0.700} & \multicolumn{1}{c}{\textbf{0.733}} & \multicolumn{1}{c}{\textbf{0.733}} & \multicolumn{1}{c}{0.700} \\
 \hline
 \multicolumn{1}{c}{\hspace{1em}-} & \multicolumn{1}{c}{0.344} & \multicolumn{1}{c}{\textbf{0.667}} & \multicolumn{1}{c}{0.333} & \multicolumn{1}{c}{0.356} \\
 \hline
 \multicolumn{1}{c}{\hspace{1em}AUC} & \multicolumn{1}{c}{0.805} & \multicolumn{1}{c}{0.155} & \multicolumn{1}{c}{0.799} & \multicolumn{1}{c}{\textbf{0.795}} \\
 \hline\addlinespace[0.5cm]
 \hline
 \multicolumn{5}{l}{\textbf{2.B}}\\
 \hline
 \multicolumn{1}{c}{\hspace{1em}+} & \multicolumn{1}{c}{0.533} & \multicolumn{1}{c}{\textbf{0.600}} & \multicolumn{1}{c}{\textbf{0.600}} & \multicolumn{1}{c}{\textbf{0.600}} \\
 \hline
 \multicolumn{1}{c}{\hspace{1em}-} & \multicolumn{1}{c}{0.489} & \multicolumn{1}{c}{\textbf{0.811}} & \multicolumn{1}{c}{0.189} & \multicolumn{1}{c}{0.211} \\
 \hline
 \multicolumn{1}{c}{\hspace{1em}AUC} & \multicolumn{1}{c}{0.537} & \multicolumn{1}{c}{0.533} & \multicolumn{1}{c}{\textbf{0.942}} & \multicolumn{1}{c}{0.938} \\
 \hline\addlinespace[0.5cm]
 \hline
 \multicolumn{5}{l}{\textbf{2.C}}\\
 \hline
 \multicolumn{1}{c}{\hspace{1em}+} & \multicolumn{1}{c}{0.540} & \multicolumn{1}{c}{0.797} & \multicolumn{1}{c}{0.793} & \multicolumn{1}{c}{\textbf{0.817}} \\
 \hline
 \multicolumn{1}{c}{\hspace{1em}-} & \multicolumn{1}{c}{\textbf{0.527}} & \multicolumn{1}{c}{\textbf{0.527}} & \multicolumn{1}{c}{0.097} & \multicolumn{1}{c}{0.103} \\
 \hline
 \multicolumn{1}{c}{\hspace{1em}AUC} & \multicolumn{1}{c}{0.525} & \multicolumn{1}{c}{0.372} & \multicolumn{1}{c}{0.834} & \multicolumn{1}{c}{\textbf{0.847}} \\
 \hline\addlinespace[0.5cm]
 \hline
 \multicolumn{5}{l}{\textbf{2.D}}\\
 \hline
 \multicolumn{1}{c}{\hspace{1em}+} & \multicolumn{1}{c}{0.917} & \multicolumn{1}{c}{\textbf{0.933}} & \multicolumn{1}{c}{\textbf{0.933}} & \multicolumn{1}{c}{0.917} \\
 \hline
 \multicolumn{1}{c}{\hspace{1em}-} & \multicolumn{1}{c}{0.050} & \multicolumn{1}{c}{\textbf{0.950}} & \multicolumn{1}{c}{0.050} & \multicolumn{1}{c}{0.067} \\
 \hline
 \multicolumn{1}{c}{\hspace{1em}AUC} & \multicolumn{1}{c}{0.747} & \multicolumn{1}{c}{0.410} & \multicolumn{1}{c}{0.752} & \multicolumn{1}{c}{\textbf{0.760}} \\
 \hline\addlinespace[0.5cm]
 \hline
 \multicolumn{5}{l}{\textbf{2.E}}\\
 \hline
 \multicolumn{1}{c}{\hspace{1em}+} & \multicolumn{1}{c}{0.624} & \multicolumn{1}{c}{\textbf{0.699}} & \multicolumn{1}{c}{0.611} & \multicolumn{1}{c}{0.617} \\
 \hline
 \multicolumn{1}{c}{\hspace{1em}-} & \multicolumn{1}{c}{0.382} & \multicolumn{1}{c}{\textbf{0.699}} & \multicolumn{1}{c}{0.368} & \multicolumn{1}{c}{0.374} \\
 \hline
 \multicolumn{1}{c}{\hspace{1em}AUC} & \multicolumn{1}{c}{0.499} & \multicolumn{1}{c}{0.530} & \multicolumn{1}{c}{\textbf{0.556}} & \multicolumn{1}{c}{0.554} \\
 \hline\addlinespace[0.5cm]
 \hline
 \multicolumn{5}{l}{\textbf{2.F}}\\
 \hline
 \multicolumn{1}{c}{\hspace{1em}+} & \multicolumn{1}{c}{0.594} & \multicolumn{1}{c}{\textbf{0.645}} & \multicolumn{1}{c}{0.611} & \multicolumn{1}{c}{0.609} \\
 \hline
 \multicolumn{1}{c}{\hspace{1em}-} & \multicolumn{1}{c}{0.387} & \multicolumn{1}{c}{\textbf{0.752}} & \multicolumn{1}{c}{0.388} & \multicolumn{1}{c}{0.391} \\
 \hline
 \multicolumn{1}{c}{\hspace{1em}AUC} & \multicolumn{1}{c}{0.501} & \multicolumn{1}{c}{0.451} & \multicolumn{1}{c}{\textbf{0.511}} & \multicolumn{1}{c}{0.501} \\
 \hline
 \end{tabu}}
 \caption{Average results on 30 simulated sampling processes. The best score in each row is indicated in bold.}\label{table_simuFI2}
 \end{table}
 
 \setlength{\tabcolsep}{6pt}

Looking at \cref{table_simuFI2}, simulation 2.A, 2.B and 2.D yield similar results to 1.A, 1.B and 1.D but with degraded performance. \textbf{Grad} performs better at correctly retrieving the sign of the effect, but its AUC is clearly lower than the ones from the other methods. On average, \textbf{Grad$\odot$Input} and \textbf{IG} seem to consistently outperform other methods to detect significant variables from the noise, notably they yield better results in simulation 2.C than in simulation 1.C. Simulation 2.E and 2.F yield mixed results. Having that many groups (83) with varying effects drasticaly decreased the AUC. \textbf{Grad$\odot$Input} and \textbf{IG} seem to be the best contenders to detect signal, but additional groups in simulation 2.E renders the methods barely better than random assignment, and additional noise in simulation 2.F renders the methods useless.

\section{Results on Spipoll data set}

\subsection{Setting}

We consider the observation period of the Spipoll data set from 2010 to 2020 included, in metropolitan France. We consider a total of $n_1=26267$ observation sessions, where $n_2=306$ taxa of insects and $u=83$ genus of plants have been observed. The observation session-insect matrix $B$ has a total of 203 244 interactions reported, and the plant-insect matrix $B'$ has 13 127 different interactions. Both BVGAE and its fair counterpart are trained on the data set, with $D_+ = D_- = 6$. For the fair-BVGAE, we  define the protected variable as the number of participations from the user at observation time. This number of participations would work as a proxy for the user's experience. By employing this measure, we hope to construct a latent space that remains unaffected by variations in observers' experience levels.

The date and place of observations allowed us to extract corresponding climatic conditions as covariates, from the European Copernicus Climate data set, and the corresponding land use proportion with the Corine Land Cover (CLC). The covariates related to the observation sessions are $X_1 = (P,t,\Delta_T,CLC)$ where $P$ is a binarized categorical variable  (83 columns) giving the plant genus, $t$ contains the day and the year of observation, $\Delta_T$ is the difference between the average temperature on the day of observation and the average of temperatures measured from 1950 to 2010 at the same observation location, and $CLC$ describes the proportion of land use with 44 categories in a 1000m radius around the observation location. To remove noise, which decreases the performance of the feature importance methods as seen in \cref{table_simuFI2}, we consider only 17 of the 44 categories, retaining those where the proportion exceeds 10\% at least 5\% of the time. 

Based on the results in \cref{table_simuFI2}, we fit BVGAE and its fair counterpart on the data set 30 times before applying \textbf{Grad$\odot$Input} and \textbf{IG} to each trained GNN to assess feature importance aggregated by plant, given by matrix $P$. The sign is then estimated using \textbf{Grad}. As $P$ is also a covariate of the model, feature importance for one plant that is aggregated with another plant's data is also estimated, these estimates are discarded for the study. Full results of feature importance using \textbf{IG} and \textbf{Grad$\odot$Input} are available on GitHub \url{https://github.com/AnakokEmre/graph_features_importance}. In the following, only results for \textbf{IG} are displayed.

\subsection{Results}

\begin{table}
  \centering
      \centering
      \begin{tabular}{|c|c|c|c|}
      \hline
      Median & Median & Features & Grad$>0$ \\
      rank & score & & \\
      \hline
      1.0 & $-3.42 \times 10^{-5}$ & Phyteuma & 0\\
      \hline
      2.0 & $-2.84 \times 10^{-5}$ & Digitalis & 0\\
      \hline
      4.0 & $+2.27 \times 10^{-5}$ & Cotoneaster & 1\\
      \hline
      6.0 & $-2.13 \times 10^{-5}$ & Valeriana & 0\\
      \hline
      6.0 & $+2.05 \times 10^{-5}$ & Verbascum & 1\\

      \hline
      \end{tabular}

  \vspace{1em} 
      \centering
      \begin{tabular}{|c|c|c|c|}
      \hline
      Median & Median & Features & Grad$>0$ \\
      rank & score & & \\
      \hline
      1.0 & $+2.49 \times 10^{-5}$ & Alliaria & 1\\
      \hline
      2.0 & $-2.05 \times 10^{-5}$ & Lavandula & 0\\
      \hline
      3.0 & $-1.88 \times 10^{-5}$ & Borago & 0\\
      \hline
      5.0 & $-1.44 \times 10^{-5}$ & Dipsacus & 0\\
      \hline
      5.5 & $-1.40 \times 10^{-5}$ & Echium & 0\\

      \hline
      \end{tabular}

  \caption{Top 5 ecological variables for connectivity according to \textbf{IG} method on 30 BGVAE (top) and Fair-BGVAE (bottom) trainings.}
  \label{tab:ecological}
\end{table}

\begin{table}
  \centering
      \centering
      \begin{tabular}{|c|c|c|c|c|}
      \hline
      Median & Median & Plant & Features & Grad $>0$ \\
      rank & score & & & \\
      \hline
      41.0 & $-8.14 \times 10^{-6}$ & Cistus & Transitional\_woodland-shrub & 0\\
      \hline
      67.0 & $-5.33 \times 10^{-6}$ & Verbascum & Complex\_cultivation\_patterns & 1\\
      \hline
      67.5 & $-4.97 \times 10^{-6}$ & Philadelphus & Discontinuous\_urban\_fabric & 0\\
      \hline
      69.0 & $+4.89 \times 10^{-6}$ & Erica & Complex\_cultivation\_patterns & 1\\
      \hline
      78.0 & $+4.81 \times 10^{-6}$ & Aquilegia & Green\_urban\_areas & 1\\
      \hline
      \end{tabular}

  \vspace{1em} 

      \centering
      \begin{tabular}{|c|c|c|c|c|}
      \hline
      Median & Median & Plant & Features & Grad $>0$ \\
      rank & score & & & \\
      \hline
      26.5 & $+6.74 \times 10^{-6}$ & Alliaria & Sport\_and\_leisure\_facilities & 1\\
      \hline
      32.0 & $+6.34 \times 10^{-6}$ & Myosotis & Mixed\_forest & 1\\
      \hline
      36.0 & $-5.92 \times 10^{-6}$ & Philadelphus & Discontinuous\_urban\_fabric & 0\\
      \hline
      41.5 & $+5.70 \times 10^{-6}$ & Phyteuma & Discontinuous\_urban\_fabric & 0\\
      \hline
      50.5 & $+5.25 \times 10^{-6}$ & Caltha & Discontinuous\_urban\_fabric & 0\\
      \hline
      \end{tabular}

  \caption{Top 5 land use variables grouped by plants for connectivity according to \textbf{IG} method on 30 BGVAE (top) and Fair-BGVAE (bottom) trainings.}
  \label{tab:landuse}
\end{table}

\begin{table}
  \centering
      \centering
      \begin{tabular}{|c|c|c|c|c|}
      \hline
      Median & Median & Plant & Features & Grad$>0$ \\
      rank & score & & & \\
      \hline
      46.5 & $+6.90 \times 10^{-6}$ & Dipsacus & Temperature & 1\\
      \hline
      54.0 & $+5.98 \times 10^{-6}$ & Verbascum & Temperature & 1\\
      \hline
      55.0 & $-5.85 \times 10^{-6}$ & Viburnum & Temperature & 1\\
      \hline
      59.0 & $-5.48 \times 10^{-6}$ & Prunus & Temperature & 1\\
      \hline
      60.0 & $+5.35 \times 10^{-6}$ & Buddleja & Temperature & 1\\
      \hline
      \end{tabular}

  \vspace{1em} 

      \centering
      \begin{tabular}{|c|c|c|c|c|}
      \hline
      Median & Median & Plant & Features & Grad$>0$ \\
      rank & score & & & \\
      \hline
      20.0 & $+7.88 \times 10^{-6}$ & Dipsacus & Temperature & 1\\
      \hline
      23.0 & $-7.32 \times 10^{-6}$ & Viburnum & Temperature & 1\\
      \hline
      28.0 & $+6.87 \times 10^{-6}$ & Verbascum & Temperature & 1\\
      \hline
      31.5 & $-6.46 \times 10^{-6}$ & Prunus & Temperature & 1\\
      \hline
      34.0 & $+6.14 \times 10^{-6}$ & Convolvulus & Temperature & 1\\
      \hline
      \end{tabular}

  \caption{Top five plants with the highest scores attributed to the 'temperature' feature for connectivity according to \textbf{IG} method on 30 BGVAE (top) and Fair-BGVAE (bottom) trainings.}
  \label{tab:temperature}
\end{table}

\begin{table}
  \centering
      \centering
      \begin{tabular}{|c|c|c|c|c|}
      \hline
      Median & Median & Plant & Features & Grad$>0$ \\
      rank & score & & & \\
      \hline
      622.0 & $-7.05 \times 10^{-7}$ & Knautia & Y & 0\\
      \hline
      961.5 & $+4.39 \times 10^{-7}$ & Lavandula & Y & 0\\
      \hline
      1121.5 & $-3.79 \times 10^{-7}$ & Plantago & Y & 0\\
      \hline
      1128.5 & $-3.72 \times 10^{-7}$ & Convolvulus & Y & 0\\
      \hline
      1222.0 & $+3.48 \times 10^{-7}$ & Carduus & Y & 0\\
      \hline
      \end{tabular}

  \vspace{1em} 

      \centering
      \begin{tabular}{|c|c|c|c|c|}
      \hline
      Median & Median & Plant & Features & Grad$>0$ \\
      rank & score & & & \\
      \hline
      1085.5 & $-3.81 \times 10^{-7}$ & Knautia & Y & 0\\
      \hline
      1569.5 & $-2.29 \times 10^{-7}$ & Plantago & Y & 0\\
      \hline
      1611.0 & $-2.18 \times 10^{-7}$ & Convolvulus & Y & 0\\
      \hline
      1615.0 & $-2.16 \times 10^{-7}$ & Erica & Y & 0\\
      \hline
      1659.0 & $+2.07 \times 10^{-7}$ & Philadelphus & Y & 0\\
      \hline
      \end{tabular}

  \caption{Top five plants with the highest scores attributed to the 'year' feature for connectivity according to \textbf{IG} method on 30 BGVAE (top) and Fair-BGVAE (bottom) trainings.}
  \label{tab:year}
\end{table}

A total of 1826 of aggregated features are estimated and ranked from most to least important. The median rank for each feature is calculated. The results presented in the following \cref{tab:ecological,tab:landuse,tab:temperature,tab:year} are sorted by median rank. ``Median score'' is the score estimated by \textbf{IG}, Grad$>0$ is the proportion of times when \textbf{Grad} has estimated a positive effect on connectivity.
 
As demonstrated in the simulation sections, we must exercise caution when interpreting the results, as it is very hard to detect signal and to estimate the sign. First, we focus on the BVGAE model. Looking at \cref{tab:temperature,tab:landuse}, it seems that both ecological and land use variables are present in the top 100 most influential variable in connectivity. However, among the top 100 most influential features, 58 of them are related to plant identity, representing 70\% of the plant genera. \textit{Green urban areas} and \textit{Discontinuous urban fabric} are influential variables on connectivity, and their effects are consistently estimated to be respectively positive and negative by the \textbf{Grad} method. A parallel could be drawn with the results of \cite{baldock_systems_2019}, who also found that gardens, parks or allotment were visited by large numbers of pollinators compared to man-made surfaces such as industrial estates. \textit{Complex cultivation patterns} are also highlighted as important variable that positively influence connectivity. This finding could be consistent with the research by \cite{deguinesWhereaboutsFlowerVisitors2012} and \cite{redhead_potential_2018}, which identified that agricultural land cover could increase pollinator generality. 28 of the top 100 most influential features are related to Temperature. All scores associated with the ``temperature'' variable (\cref{tab:temperature}) are estimated as positive by \textbf{Grad}, while all scores associated with the ``year'' variable are estimated as negative. This result could be similar to the ones of \cite{duchenne_longterm_2020}, who also showed that some bees species benefited from temperature increase, with an overall decline in insect occupancy over the years.

If we focus on the fair-BVGAE model (\cref{tab:ecological}), \textbf{IG} and \textbf{Grad$\odot$Input} identify the genera \textit{Alliaria}, \textit{Borago}, \textit{Lavandula}, and \textit{Dipsacus} as the most important variables. Among the top 100 most influential features, 44 of them are related to plant identity. Accounting for observers' experience levels has led to a higher ranking for the variable ``temperature'', as 42 of the top 100 most influential features are related to it. The scores associated with ``temperature'' and ``year'' (\cref{tab:temperature,tab:year}) variables have the same sign as those estimated in BVGAE. Compared to the BVGAE, the sign of \textit{Green urban areas} has changed to the opposite. While connectivity may appear positively influenced in the BVGAE due to the presence of green areas in urban environments, which attract more visitors, the effect may be more nuanced. When the user experience is taken into account, the connectivity could actually be negatively impacted compared to other locations, as the area remains within an urban setting. This hypothesis is difficult to assert because, as seen in the simulation, the estimated sign has to be interpreted with caution. 

The complete ranking estimated by \textbf{IG} on BVGAE and fair-BVGAE are overall correlated, as they present a Spearman correlation of 0.73. Among the top 100 most important variables of BVGAE, 63 are also estimated to be the top 100 most important variables of fair-BVGAE, 32 of which are solely related to plant genera. The Spearman correlation for the top 100 most important variables drops to 0.21, indicating that even if plants are still overall estimated as important, their order of importance changes when the method takes into account the sampling effects. This could suggest that sampling effects bias observation differently depending on the plant species that is monitored. For land use variables aggregated by plants, \textit{Green urban areas} and \textit{Discontinuous urban fabric} both appear 3 times on BVGAE top 100 rankings, associated respectively with the genera \textit{Aquilegia}, \textit{Chelidonium}, \textit{Sedum}, and \textit{Philadelphus}, \textit{Caltha}, \textit{Cotoneaster}. In fair-BVGAE, \textit{Discontinuous urban fabric} remains in the top 100, linked with the previously mentionned genera, \textit{Phyteuma} and \textit{Scabiosa}. However, the scores associated with \textit{Green urban areas} drop and this feature does not appear in the top 100 anymore, as the first occurence is at median rank 247. These results can highlight how accounting for observers' experience levels could potentially change the perception of which features influence the plant-pollinator network connectivity.

\correction{The methodology can provide interesting insights of the impact of land use and global warming on plant-pollinator network connectivity, even if we must exercise extensive caution when interpreting the results. The estimated scores of flower genera were greater than the ones for land uses or temperature, suggesting that the identity of plant species might matter more than the surrounding landscape in determining network connectivity. Such  effect of plant identity on network connectivity is coherent with studies suggesting relationships between plant traits and plant generalism \citep{Ollerton, Galetto}
Our results also suggest that effects of global change drivers on network connectivity differ among the plant species. This further suggest that global change drivers may impact plants genera differently, as we know that some species benefit from global change while others are negatively affected \citep{Timmermann_Pervasive_2015,duchenne_longterm_2020}.
Most importantly, the estimation of these effects may be influenced by the observers' experience levels. This highlights the need for caution when dealing with citizen science data, as the estimated effects on connectivity can be biased by the observers.
Finally, using similar approaches, further studies could investigate the effect of conservation action on plant-pollinator communities.}

\section{Conclusion}
In this work, we explored interpretability methods for GNNs to assess the impact of node features on network connectivity \correction{through an extensive simulation study where the data acquisition process was modeled}. Features with only one effect can be efficiently detected, and their sign can be estimated.  However, if the effect depends on groups of nodes (e.g. plant genera), then all presented methods may struggle to retrieve the sign of the effect and its magnitude. 
Then, we assess what impact connectivity in the pollination network issued from the spipoll dataset.

\correction{The simulation processes described in Section \ref{sec:simstu}, which respectively reproduce the generation of a bipartite network (Section \ref{settings_on_simulated_bipartite_networks}) and the sampling procedure of the Spipoll dataset (Section \ref{setsimulatedsamplingprocess}), play a crucial role in evaluating the performance of statistical learning methods on such data. In our work, we focused on assessing the accuracy of interpretability methods developed for GNNs. 
These simulation frameworks (particularly the second one, which models a sampling process affected by biases) could also be valuable for other studies conducted in similar contexts, aiming to examine the impact of such biases on the accuracy of statistical learning approaches.
Although designing the simulation model was not the main objective of this paper, particular care was taken to ensure its flexibility and capacity to represent a range of scenarios.
The development of such models should follow general good modeling practices \citep{jakeman2024towards}, while maintaining a specific emphasis on their alignment with the statistical methods to be evaluated on the synthetic data generated through them.}

\correction{Despite their limitations, interpretability methods can provide valuable insights on bipartite network data. The data at hand in this paper are large since they are issued from a citizen science program. Data from scientific campaigns are smaller which limits the training of a GNN that needs  relatively large datasets.
 However, numerous collections of these networks \citep{dore_relative_2021, fortuna2014weblife} were built. Therefore, the methodology we propose could be applied on such collections, although some adaptations could be necessary to handle the nature of these data which are a collection and not a single network. We focused on connectivity but other metrics (nestedness, modularity \citealt{BascompteNestedness2003,olesen2007modularity}) describing the structure of a network may be explored as well. Co-occurence data could also be dealt with as bipartite network \citep{Bernardo,Bernardo2}. In this case, our approach could be used to assess the impact on the modularity score of the different site covariates. }

\section{Declaration of generative AI and AI-assisted technologies in the writing process.}

During the preparation of this work the authors used ChatGPT to improve fluidity in English. After using this tool, the authors reviewed and edited the content as needed and take full responsibility for the content of the published article.

\bibliography{biblio.bib}

\end{document}